\pdfoutput=1

\documentclass[sigconf,nonacm]{acmart}

\usepackage{booktabs} 

\usepackage{multirow}

\setcopyright{none}

\begin{document}
\title{LayGA: Layered Gaussian Avatars for Animatable Clothing Transfer}

\author{Siyou Lin}
\email{linsy21@mails.tsinghua.edu.cn}
\author{Zhe Li}
\email{liz19@mails.tsinghua.edu.cn}
\author{Zhaoqi Su}
\email{suzq13@tsinghua.org.cn}
\affiliation{%
 \institution{Tsinghua University}
 \streetaddress{30 Shuangqing Road}
 \city{Haidian}
 \state{Beijing}
 \postcode{100084}
 \country{China}}
\author{Zerong Zheng}
\affiliation{%
 \institution{NNKosmos Technology}
 \city{Hangzhou}
 \state{Zhejiang}
 \country{China}
}
\email{zrzheng1995@foxmail.com}

\author{Hongwen Zhang}
\affiliation{%
 \institution{Beijing Normal University}
 \streetaddress{19 Xinjiekouwaidajie Street}
 \city{Haidian}
 \state{Beijing}
 \postcode{100875}
 \country{China}}
\email{zhanghongwen@bnu.edu.cn}

\author{Yebin Liu}
\affiliation{%
 \institution{Tsinghua University}
 \streetaddress{30 Shuangqing Road}
 \city{Haidian}
 \state{Beijing}
 \postcode{100084}
 \country{China}
}
\authornote{Corresponding author}

\renewcommand\shortauthors{Lin, S. et al}

\begin{abstract}
Animatable clothing transfer, aiming at dressing and animating garments across characters, is a challenging problem.
Most human avatar works entangle the representations of the human body and clothing together, which leads to difficulties for virtual try-on across identities.
What's worse, the entangled representations usually fail to exactly track the sliding motion of garments.
To overcome these limitations, we present Layered Gaussian Avatars (LayGA), a new representation that formulates body and clothing as two separate layers for photorealistic animatable clothing transfer from multi-view videos. 
Our representation is built upon the Gaussian map-based avatar for its excellent representation power of garment details.
However, the Gaussian map produces unstructured 3D Gaussians distributed around the actual surface.
The absence of a smooth explicit surface raises challenges in accurate garment tracking and collision handling between body and garments.
Therefore, we propose two-stage training involving single-layer reconstruction and multi-layer fitting.
In the single-layer reconstruction stage, we propose a series of geometric constraints to reconstruct smooth surfaces and simultaneously obtain the segmentation between body and clothing.
Next, in the multi-layer fitting stage, we train two separate models to represent body and clothing and utilize the reconstructed clothing geometries as 3D supervision for more accurate garment tracking.
Furthermore, we propose geometry and rendering layers for both high-quality geometric reconstruction and high-fidelity rendering.
Overall, the proposed LayGA realizes photorealistic animations and virtual try-on, and outperforms other baseline methods. Our project page is \href{https://jsnln.github.io/layga/index.html}{https://jsnln.github.io/layga/index.html}.
\end{abstract}

%

\ccsdesc[500]{Computing methodologies~Shape modeling}

%
%

\keywords{Animatable avatar, clothing transfer, human reconstruction}

\begin{teaserfigure}
    \centering
    \includegraphics[width=\linewidth]{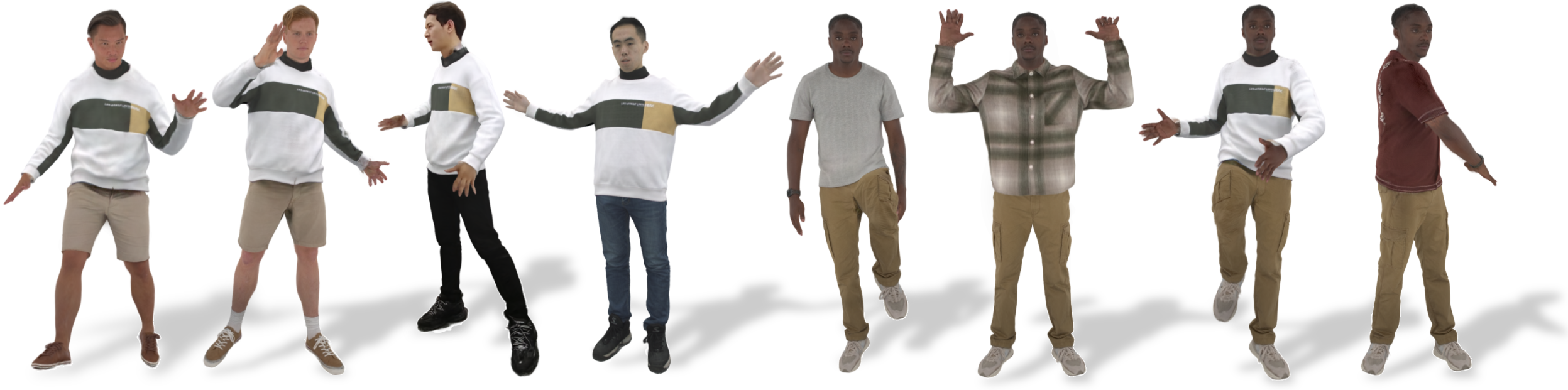}
    \caption{Our method can create layered animatable avatars for clothing transfer with realistic garment details. The left shows four characters wearing the same upper garment, while the right shows four different garments are dressed on the same character.}
    \label{fig:teaser}
\end{teaserfigure}

\maketitle

\section{Introduction}

Photorealistic human avatars have drawn considerable attention with recent advances in rendering techniques, which aim to create a photo-realistic virtual digital embodiment of a clothed human.
While Neural Radiance Fields (NeRFs)~\cite{mildenhall2020nerf} have previously dominated the area of photo-realistic avatar creation~\cite{peng2022animatable,su2021a-nerf,zheng2022structured,Feng2022scarf,weng2022humannerf,te2022neural,peng2022selfnerf,guo2023vid2avatar,jiang2023instantavatar,jiang2022neuman,li2022tava,wang2022arah,li2023posevocab,zheng2023avatarrex}, the NeRF-based avatar methods still face a challenge in accurately representing high-frequency human dynamics.
Additionally, the rendering speed for NeRF is relatively slow.
There is an ongoing trend to replace NeRFs with the recently proposed 3D Gaussian Splatting (3DGS)~\cite{kerbl2023gaussian} in human avatars \cite{li2023animatable,zhu2023ash,Zielonka2023Drivable3D,hu2023gauhuman,kocabas2023hugs}. 
3DGS, being both differentiable and highly efficient, has become the new go-to representation for photorealistic novel-view synthesis. 
However, most existing works that apply NeRF or 3DGS to clothed humans consider body and clothing as a unified single layer. 
Albeit adequate for most cases, this formulation encounters challenges in modeling sliding motions between different garment layers, and cannot accommodate certain applications such as clothing transfer.

Compared with single-layer representations, multi-layer modeling allows more accurate tracking of the sliding movement of clothing boundaries~\cite{xiang2021modeling,yu2019simulcap}. 
Furthermore, layered modeling also enables new applications such as virtual try-on by transferring the clothing of one character to the body of another~\cite{pons2017clothcap}. 
However, there are few NeRF or 3DGS works that support the modeling of multi-layer human avatars.
SCARF~\cite{Feng2022scarf} and DELTA~\cite{feng2023learning} propose to combine NeRF-based clothing and mesh-based body, which are capable of representing different geometric properties of clothing and body layers and support clothing transfer. 
However, the rendering quality and garment dynamics do not fully meet the desired level, which may be attributed to NeRF's limited capability for thin-layered clothing. 

In this work, we propose Layered Gaussian Avatars (LayGA), which model animatable multi-layered clothed humans using Gaussian splats, seamlessly integrating state-of-the-art 3DGS-based human avatars with the ability for clothing transfer. 
3DGS is chosen for its advantages in high-fidelity and efficient rendering.
The explicit representation provides a clearer depiction of geometric details in thin-layered clothing compared to NeRF. 
Our layered avatar representation is built upon the Gaussian map-based avatar \cite{li2023animatable}, which learns pose-dependent 3D Gaussians on the 2D domain, because of its ability of modeling high-frequency clothing dynamics.
However, naively learning two sets of Gaussians to separately represent body and clothing is infeasible. 
First, 3D Gaussians optimized through differentiable rendering without any constraints are typically unevenly distributed around the actual human surface. 
Consequently, they fail to provide an explicit and smooth geometric surface for modeling the body-cloth  relations, which is essential for adapting clothing to different body shapes and collision handling. 
Secondly, while parametric body models like SMPL~\cite{loper2015smpl}/SMPL-X~\cite{SMPL-X:2019} can be employed for the body part as a prior, the appearance of the clothing template is not predetermined in advance.
Concurrent work D3GA~\cite{Zielonka2023Drivable3D} proposes a multi-layer human avatar representation using 3D Gaussian Splatting. However, the synthesized avatar is blurry due to the limited capability of their MLP-based representation, and it currently lacks support for clothing transfer.

To overcome the above challenges, we propose two-stage training involving single-layer reconstruction and multi-layer fitting.
In the single-layer reconstruction stage (Sec.~\ref{subsec:single-layer}), we introduce a series of geometrical constraints to force the 3D Gaussians to lie on a smooth surface, thus providing explicit geometries for collision handling in the following layered modeling.
Besides, we learn a segmentation label channel to separate the clothing from the unified 3D Gaussians.
In the next multi-layer fitting stage (Sec.~\ref{subsec:multi-layer}), we train two separate models to represent the body and clothing.
The previously reconstructed clothing geometry serves as a 3D proxy for more accurate tracking of clothing motion.
Moreover, we propose an additional rendering layer to guarantee both smooth reconstruction and high-fidelity rendering, since smooth surfaces obtained by the proposed geometric constraints may degenerate the rendering quality.
Overall, once the training stages finish, our layered model is able to not only generate realistic animation under novel poses, but also transfer the clothing across identities.

To summarize, our contributions include:

\begin{itemize}
  \item We propose layered Gaussian Avatars (LayGA), the first 3DGS-based layered human avatar representation for animatable clothing transfer.
  \item We introduce geometric constraints on 3D Gaussians for smooth surface reconstruction, supporting collision handling between the body and clothing in the layered representation.
  \item In the multi-layer learning, we introduce previously segmented reconstruction as supervisions for more accurate tracking of clothing boundaries. We additionally introduce a rendering layer to alleviate the deterioration of rendering quality brought by geometric constraints.
\end{itemize}

\begin{figure*}
    \centering
    \includegraphics[width=\linewidth]{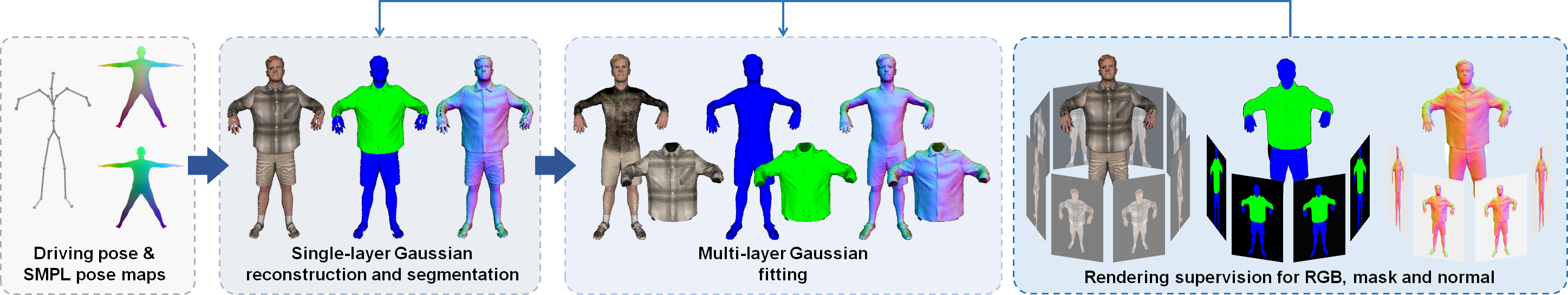}
    \caption{Overview of our pipeline. Our pipeline consists of two training stages: 1) single-layer reconstruction and segmentation; 2) Multi-layer fitting.}
    \label{fig:pipeline}
\end{figure*}

\section{Related Work}

\subsection{Integrated Modeling of Clothed Human Avatars}
Human avatar aims to formulate realistic pose-dependent human and clothing motions under novel poses. Most human avatars model clothing and the body as a whole. Traditional mesh-based human avatars~\cite{ma2020learning,burov2021dynamic,saito2021scanimate,habermann2021real,habermann2023hdhumans,NPMS} formulate human avatar with a single fixed-topology human mesh, aims to explicitly represent the pose-dependent human geometry as a deformation field over the mesh vertices. However, these methods rely on a pre-defined human mesh, which can hardly represent complicated clothing topologies or detailed pose-dependent geometry and texture. 
Recently, to represent more flexible geometry and garment movements, controllable human avatars mainly rely on point-based or Neural Radiance Field~\cite{mildenhall2020nerf} (NeRF)-based methods. 
For point-based methods, POP~\cite{ma2021power} maps the semantic human point cloud onto the SMPL UV space and learns the pose-dependent UV features for human avatar animation. Furthermore, FITE~\cite{lin2022learning} and CloSET~\cite{zhang2023closet} are proposed to flexibly learn the point-cloud template for humans under loose garments. DPF~\cite{prokudin2023dynamic} solves an as-isometric-as-possible deformation field for better representing complicated garment motions. Although these methods can be applied to various garment types and motions, point-based methods face limitations when applied to RGB-based inputs due to their lack of textual information.

As an implicit neural rendering technique with dense geometry and texture field, NeRF-based human avatars~\cite{peng2021neural,liu2021neural,peng2021animatable,su2021a-nerf,hnerf,weng2022humannerf,peng2022selfnerf,hu2023sherf,li2022tava,zheng2022structured,zheng2023avatarrex,li2023posevocab,jiang2023instantavatar} are capable of learning view-dependent textual details from multi-view or monocular RGB inputs. 
To model better dynamics, AnimatableNeRF~\cite{peng2021animatable} defines a NeRF deformation field over the canonical SMPL~\cite{loper2015smpl}; SLRF~\cite{zheng2022structured} defines local NeRF fields on the sampled human nodes;
PoseVocab~\cite{li2023posevocab} employs a pose vocabulary for encoding high-frequency local dynamic details. Efficiency-wise, InstantAvatar~\cite{jiang2023instantavatar} achieves one-minute training for an avatar from a monocular human video.
Although these methods achieve plausible quality, NeRF-based methods still lead to over-smooth results especially on high-frequency textual regions, due to the low-frequency bias of MLPs. Besides, their rendering speed is relatively slow due to dense sampling.

Gaussians are another popular choice for representing humans. Early works have adopted Gaussians associated to skeletons for human/hand pose tracking~\citep{stoll2011sog,sridhar2014handsag,sridhar2015hand,rhodin2015versatile,rhodin2015volcontour}, but do not attempt to modeling photo-realistic appearance. Recently, with the emergence of 3D Gaussian Splatting (3DGS)~\cite{kerbl2023gaussian}, which achieves SOTA rendering speed and quality, Gaussian-based representations become the new go-to choice for high-quality rendering of human avatars~\cite{li2023animatable,kocabas2023hugs,moreau2023human,ye2023animatable,zhu2023ash}. 3D Gaussian Splatting combines the advantages of explicit point-based modeling for efficiently representing flexible geometry, and the neural rendering technique for learning pose-dependent textual information from RGB inputs like NeRF. With the StyleUNet-encoded Gaussian parameter learning technique and self-adaptive template learning, AnimatableGaussians~\cite{li2023animatable} achieves highly dynamic, realistic and generalized details for human avatar animation. However, the above methods all model the human and clothing as a whole, which hinders their applicability to specific use cases, including but not limited to clothing transfer.

\subsection{Layered Modeling of Clothed Human Avatars}

To better represent garment properties on top of the human bodies and apply to garment transfer or editing, some researchers propose disentangled human avatars, which model garments as separate layers. Traditional layered-based clothed human avatars are mesh-based methods. Some methods leverage a dense multi-view capture system to reconstruct high-quality human and garment meshes with temporal consistency, and learn pose-dependent human and garment texture maps for clothed avatar animation~\cite{bagautdinov2021driving,xiang2021modeling,xiang2022dressing}. CaPhy~\cite{caphy_su2023} learns garment dynamics from clothed human scans and unsupervised physical energies. DiffAvatar~\cite{li2023diffavatar} constructs a human avatar from a single scan, which jointly solves the 2D clothing pattern, clothing material properties and human shape, and performs physical-based human avatar driving. These methods either rely on expensive capture systems~\cite{bagautdinov2021driving,xiang2021modeling,xiang2022dressing} or can hardly learn pose-dependent garment texture and geometry from RGB inputs~\cite{caphy_su2023,li2023diffavatar}.

Recently, a few works focus on layered modeling of clothed human avatars with NeRF~\cite{mildenhall2020nerf} or mesh surfaces to learn garment properties from RGB inputs. SCARF~\cite{Feng2022scarf} and DELTA~\cite{feng2023learning} combine implicit NeRF-based garment and explicit mesh-based body modeling to better represent each individual part. GALA~\citep{kim2024gala} adopts DMTet~\citep{shen2021dmtet} to represent different layers of clothed humans but focuses on generation conditioned on a single scan. While these methods enables clothing transfer to different bodies, the reconstructed garment texture still lacks high-frequency and dynamic details. Concurrent work D3GA~\cite{Zielonka2023Drivable3D} uses Gaussian Splatting for modeling layered humans but do not focus on garment transfer.
\section{Method}

Our model is a pose-conditioned generator of layered 3D Gaussians that produces photorealistic animations of human avatars and enables clothing transfer across identities. 
Fig.~\ref{fig:pipeline} shows our main pipeline. 
Given a body pose, we first convert it into a position map as the pose condition, and then use a StyleUNet-based~\citep{wang2023styleavatar} model to predict pose-dependent 3D Gaussians.
These 3D Gaussians are non-rigidly deformed from a SMPL-X \cite{SMPL-X:2019} template in the canonical space, posed using linear blend skinning (LBS), and subsequently rendered to the given view by 3DGS~\citep{kerbl2023gaussian}.
We empirically find that pure photometric cues are insufficient for tracking garments motions, so we propose to divide the training procedure into two stages: ({\romannumeral 1}) single-layer reconstruction and segmentation; ({\romannumeral 2}) multi-layer fitting.

In the single-layer reconstruction stage, we obtain segmented reconstruction by employing our proposed geometric constraints and using garment mask supervision~\cite{li2020self}. In the multi-layer fitting stage, we train two layers of Gaussians (body and clothing)\footnote{In our current setting, we only consider upper clothes as the clothing part. However, the formulation can be extended to any outmost-layer garment(s).} using previously obtained segmented geometries. Once trained, our model not only generates photorealistic avatar animations, but also enables clothing transfer across different characters.

\subsection{Clothing-aware Avatar Representation}

\begin{figure}[t]
    \centering
    \includegraphics[width=\linewidth]{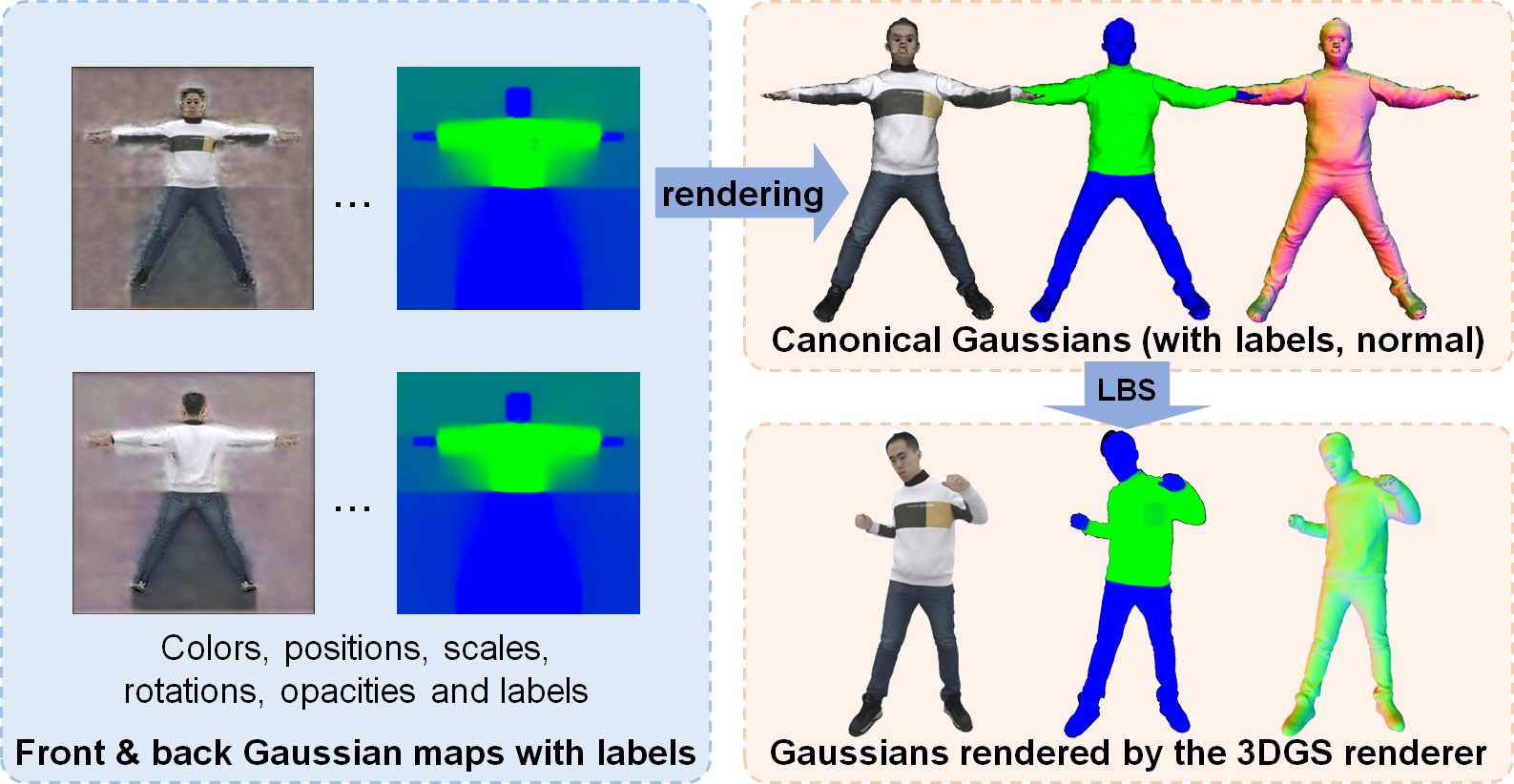}
    \caption{Illustration of the clothing-aware avatar representation.}
    \label{fig:model}
\end{figure}

As illustrated in Fig.~\ref{fig:model}, our model is built upon a recent state-of-the-art 3DGS avatar representation, Animatable Gaussians~\cite{li2023animatable}, which predicts pose-dependent Gaussian maps in the 2D domain for modeling higher-fidelity human dynamics.
Our clothing-aware model adopts a similar architecture with \citet{li2023animatable} but introduces modifications tailored to our layered modeling.

Given a body pose $\theta$ represented as the SMPL-X joint angles, we first convert it to a position map $M_{\rm pos}\in\mathbb{R}^{H\times W\times 6}$ as a conditioning signal (front and back maps concatenated channel-wise, with $H=W=512$). The position map is obtained by rendering the canonical SMPL-X model to the front and back views, with vertices colored using the posed coordinates. 
As demonstrated in \cite{li2023animatable}, such a 2D parameterization allows us to utilize powerful 2D convolutional networks (CNN), e.g., StyleUNet \cite{wang2023styleavatar}, to predict Gaussian parameters for modeling higher-fidelity appearances.
The position map $M_{\rm pos}$ is then fed to a StyleUNet-based model to generate front and back Gaussian maps $M_{\rm g}^{\rm f}, M_{\rm g}^{\rm b}\in\mathbb{R}^{H\times W\times C}$, where $C=16$.
Different from the method of \citet{li2023animatable} that only predicts Gaussian parameters including color (3-dim), offset/position (3-dim), opacity (1-dim) and covariance (7-dim), our clothing-aware model additionally learns a label, represented as the probability that each 3D Gaussian belongs to body or clothing.

Given the predicted Gaussian maps $M_{\rm g}^{\rm f}$ and $M_{\rm g}^{\rm b}$, we can extract 3D Gaussians inside the template mask.
Let us denote the Gaussian parameters of each 3D Gaussian as: color $c_i\in\mathbb{R}^3$, position offset from SMPL $\Delta \bar x_i\in\mathbb{R}^3$, opacity $o_i\in\mathbb{R}$, scales $\bar s_i\in\mathbb{R}^3$, rotations $\bar q_i\in\mathbb{R}^4$ and probabilities $p^{\rm cloth}_i$ and $p^{\rm body}_i$ (normalized by softmax). 
Here, the superscript bar in $\Delta \bar x_i$ $\bar s_i$ and $\bar q_i$ means these predictions are defined in the canonical space. 
Let $\bar x^{\rm smpl}_i$ denote the point on SMPL corresponding to pixel $i$. 
Then the mean and covariance of the 3D Gaussian are formulated as
\begin{eqnarray}
    \bar x_i&=&\bar x^{\rm smpl}_i+\Delta\bar x_i,\\
    x_i&=&R_i(\theta)\bar x_i+t_i(\theta),\label{eq:gaussposition}\\
    \Sigma_i&=&R_i(\theta)\bar\Sigma_i R_i(\theta)^T.
\end{eqnarray}
Here, $R(\theta)$ and $t(\theta)$ represent the LBS transformation given the driving pose $\theta$, and $\bar\Sigma_i$ is the covariance matrix in the canonical space derived from $\bar s_i$ and $\bar q_i$.
The posed 3D Gaussians are eventually rendered to an image using the 3DGS renderer~\citep{kerbl2023gaussian}.

\subsection{Single-Layer Modeling with Geometric Constraints}
\label{subsec:single-layer}

In the single-layer reconstruction stage, our goal is to train a model that produces 3D Gaussians smoothly distributed on the actual geometric surface of the captured human, and simultaneously obtain the segmentation between body and clothing.

\subsubsection{Geometric Constraints for Reconstruction}
As the vanilla 3DGS does not involve any geometric constraints in the training procedure, the resulting 3D Gaussians will not converge to a smooth surface but are disorderly scattered around the actual surface.
The reconstructed surfaces of clothed humans are desired to be continuous and smooth, with garment details and clear clothing boundaries.
Our key observation is that in our representation the 3D Gaussians correspond to evenly space pixels on a 2D map, so we can conveniently constrain the underlying geometry using the neighborhood of each pixel.
Specifically, we additionally introduce the following geometric constraints for regularization and detail enhancement.

\begin{figure}
    \centering
    \includegraphics[width=0.8\linewidth]{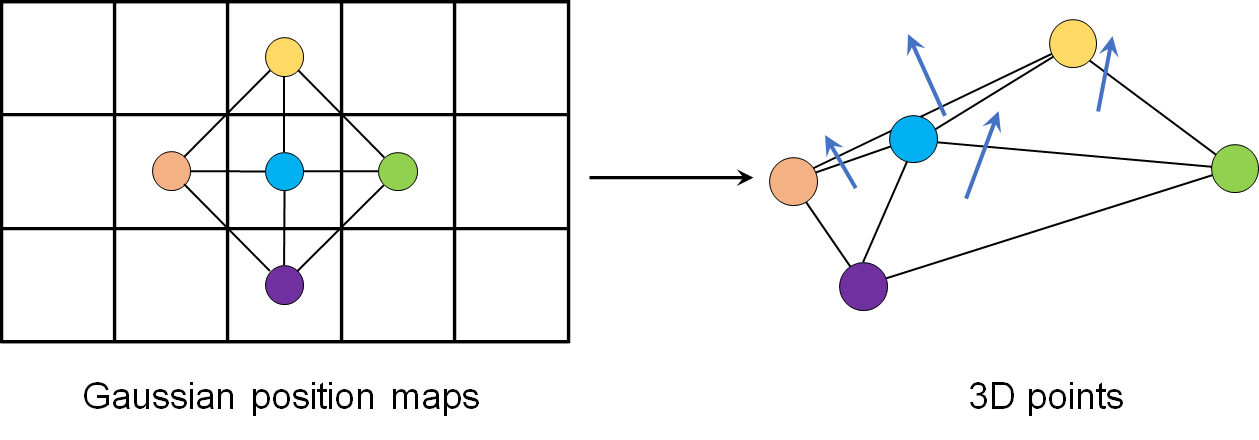}
    \caption{Illustration of normal computation on the Gaussian map.}
    \label{fig:normalcompute}
\end{figure}

\paragraph{Image-based Normal Loss} A main difficulty for multi-view reconstruction using differentiable rendering is the shape-radiance ambiguity, i.e., a wrongly reconstructed geometry may produce correct rendering for training views. To solve the ambiguity, we propose to use normals estimated from images as an additional supervision signal. Fortunately, recent works allow normals to be relatively accurately estimated for clothed humans~\citep{saito2020pifuhd,xiu2023econ}. We derive normals from our model by utilizing image pixel neighbors. 
As shown in Fig.~\ref{fig:normalcompute}, for each pixel $i$, we use the triangles formed by $i$ and its neighbors to compute its normals. As an example, suppose the neighbors of $i$ are $j,k,l,m$ (arranged counter-clockwise). Then the normal $n_i$ on this pixel is:
\begin{eqnarray}
    n_i&=&R_i(\theta)\bar n_i/\|R_i(\theta)\bar n_i\|_2,\quad\bar n_i=\hat n_i/\|{\hat n_i}\|_2,\\
    \hat n_i&=&(\bar x_j-\bar x_i)\times(\bar x_k-\bar x_i)+(\bar x_k-\bar x_i)\times(\bar x_l-\bar x_i)\nonumber\\
    &&+(\bar x_l-\bar x_i)\times(\bar x_m-\bar x_i)+(\bar x_m-\bar x_i)\times(\bar x_j-\bar x_i).
\end{eqnarray}
When computing the normals, we only take account of pixels whose neighbors are all inside the template mask.
The normals are rendered as additional channels by rasterization, and compared with the predicted ones from color images. The normal loss $\mathcal L_{\rm normal}$ is the $L_1$ loss between the estimated normal image and the rendered normal image (averaged over all pixels).

\paragraph{Stitching Loss} Since the 3D Gaussians are parameterized on two separate maps (front and back), we introduce $\mathcal L_{\rm stitch}$, an $L_2$ loss between the boundary pixels in the front map with their counterparts in the back map to prevent discontinuity.

\paragraph{Regularization} In addition to normal supervision, we also utilize the following geometric regularization losses to penalize large distortions and make the model favor small-deformation solutions.
Following \cite{li2023animatable}, we penalize large offsets by adding an offset regularization loss $\mathcal L_{\rm off}=\frac{1}{N}\sum_{i}\|{\Delta\bar x_i}\|_2^2$.
We also use a total variational (TV) loss $\mathcal L_{\rm TV}$, which is defined as the averaged $L_2$ loss on the positional differences of all neighboring pixels. This constrains the positions of two neighboring pixels to remain close, penalizing a scattered distribution of the 3D Gaussians. 
Following \citet{pons2017clothcap}, we regularize the edge lengths between the base SMPL-X model and the deformed one using an edge regularization loss $\mathcal L_{\rm edge}$. Here, an edge is defined between two neighboring valid pixels. $\mathcal L_{\rm edge}$ is the averaged $L_2$ loss on the differences between the lengths of an edge before and after adding the offset.
These losses are summed up as the regularization term:
\begin{eqnarray}
\mathcal L_{\rm reg} = \lambda_{\rm off}\mathcal L_{\rm off}+\lambda_{\rm TV}\mathcal L_{\rm TV}\nonumber+\lambda_{\rm edge}\mathcal L_{\rm edge}.
\end{eqnarray}

Eventually, the geometric loss functions are formulated as
\begin{eqnarray}
\mathcal L_{\rm geom} = \lambda_{\rm normal}\mathcal L_{\rm normal}+\lambda_{\rm stitch}\mathcal L_{\rm stitch}+\mathcal L_{\rm reg}.
\end{eqnarray}

\subsubsection{Clothing Segmentation}
During the single-layer reconstruction stage, we also aim to obtain segmentation between body and clothing. 
Recall that we predict probabilities $p_i^{\rm body}$ and $p_i^{\rm cloth}$ (normalized by softmax) whether a given Gaussian is body or clothing. We also render these two values as additional channels using the 3DGS renderer. This gives a rendered segmentation image $S$ (2 channels, $S^{\rm body}$ and $S^{\rm cloth}$). The label loss $\mathcal L_{\rm label}$ is the cross entropy loss between the rendered segmentation image $S$ and the ground truth segmentation $S_{\rm gt}$:
\begin{eqnarray}
    \mathcal L_{\rm label}&=&-\frac{1}{N_{\rm body}}\sum_i\log(S^{\rm body}_i)-\frac{1}{N_{\rm cloth}}\sum_{i'}\log(S^{\rm cloth}_{i'})\nonumber\\&&-\frac{1}{N_{\rm bg}}\sum_{i''}\log(1-S^{\rm body}_{i''}-S^{\rm cloth}_{i''}).
\end{eqnarray}
Here, $i$ ranges over all pixels in $S_{\rm gt}$ that are segmented as body, $i'$ ranges over pixels segmented as clothing, and $i''$ ranges over all background pixels. $N_{\rm body}$, $N_{\rm cloth}$ and $N_{\rm bg}$ denote the number of pixels of the corresponding type. $S_{\rm gt}$ is obtained by considering both SCHP \cite{li2020self} masks and the matting masks that come with the dataset. 
The details can be found in the supplementary document.
Furthermore, similar to geometric constraints, we also apply an $L_1$ TV loss $\mathcal L_{\rm TV}^\text{label}$ on the predicted Gaussian label map, and a stitching loss $\mathcal L_{\rm stitch}^\text{label}$ on the labels of boundary pixels. The segmentation loss functional is:
\begin{equation}
    \mathcal L_{\rm seg}=\lambda_{\rm label}\mathcal L_{\rm label}+\lambda_{\rm TV}^\text{label}\mathcal L_{\rm TV}^\text{label}+\lambda_{\rm stitch}^\text{label}\mathcal L_{\rm stitch}^\text{label}.
\end{equation}

The rendering loss function includes an $L_1$ loss, an SSIM \cite{wang2004image} loss and a perceptual \cite{zhang2018unreasonable} loss on the rendered RGB images:
\begin{equation}
    \mathcal L_{\rm render}=\lambda_{\rm L1}\mathcal L_{\rm L1}+\lambda_{\rm ssim}\mathcal L_{\rm ssim}+\lambda_{\rm lpips}\mathcal L_{\rm lpips}.
\end{equation}
The final loss function is the sum of all three parts:
\begin{equation}
    \mathcal L=\mathcal L_{\rm render}+\mathcal L_{\rm geom}+\mathcal L_{\rm seg}.
\end{equation}
After training, we can obtain high-quality reconstruction and segmentation of body and clothing as shown in Fig.~\ref{fig:pipeline}.

\subsection{Avatar Fitting with Multi-Layer Gaussian}
\label{subsec:multi-layer}

The goal of the multi-layer modeling stage is to use the segmented geometry from the single-layer stage to build a two-layer avatar. 
Recall that the single-layer stage assigns a body-or-clothing label to each Gaussian. 
We use the labels from the first frame of each sequence (assumed to be an A-pose) to specify a subset of Gaussians that is labeled as clothing. The criterion for classifying a Gaussian as clothing is $p^{\rm cloth}>0.5$, and all other Gaussians are classified as body.
This subset will be used as a template for the clothing part, while the body part is still defined with SMPL-X.

In this stage, we train two separate models, one for body and the other for clothing. 
The two models use the same network architectures but different network weights, and the clothing model will only output the clothing subset specified above.
The model architecture remains mostly the same as the single-layer stage, but we propose the following modifications for both reconstruction and rendering quality.

\begin{figure}[t]
    \centering
    \includegraphics[width=\linewidth]{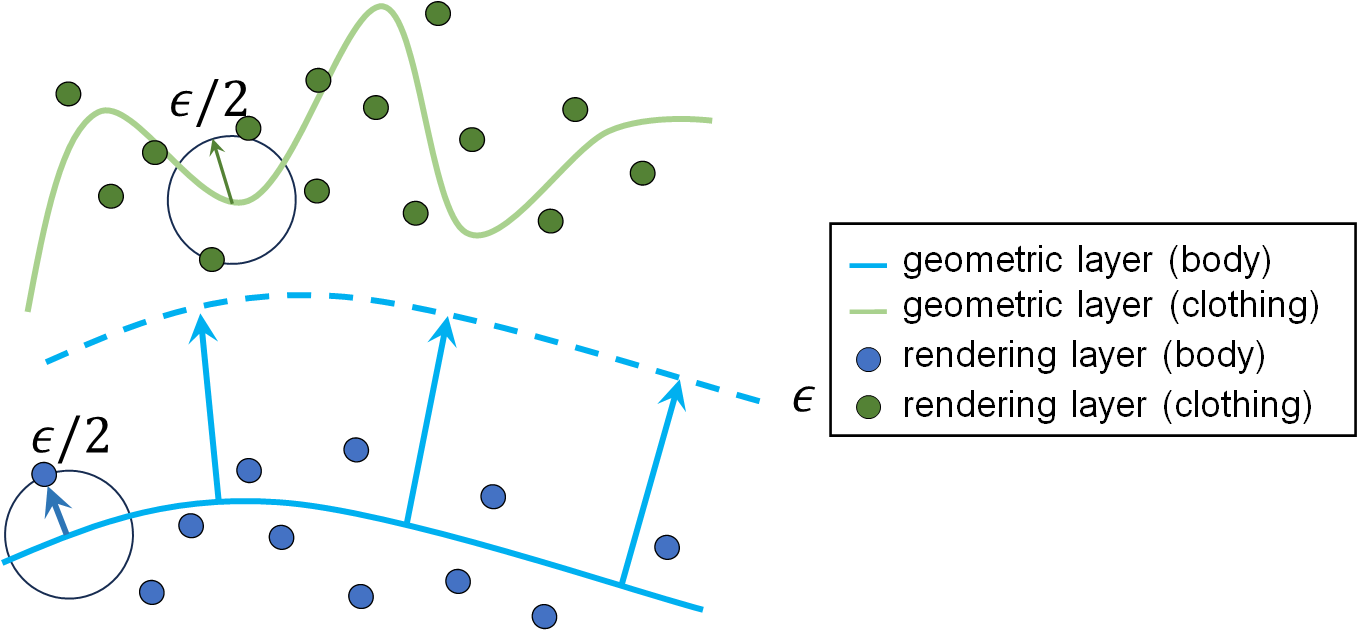}
    \caption{Illustration of geometric and rendering layers. $\epsilon$ is the threshold for handling collisions.}
    \label{fig:geomrenderlayers}
\end{figure}

\subsubsection{Separating Geometry and Rendering Layers}
As described in Sec.~\ref{subsec:single-layer}, geometric constraints force the 3D Gaussians to converge to a smooth surface. 
However, we empirically found that these geometric constraints negatively impacted the rendering quality of 3DGS, possibly because these geometric constraints lower the flexibility of Gaussians to model high-fidelity appearance. 
To prevent the adverse impact brought by geometric constraints, while still preserving a smooth geometry for collision handling in clothing transfer, we propose to separate the geometry layer and the rendering layer. 
As shown in Fig.~\ref{fig:geomrenderlayers}, in addition to $\Delta\bar x_i$, we add a second offset $\Delta\bar y_i$. 
The Gaussian positions with only the first offset computed as in Eq.~\eqref{eq:gaussposition} are referred to as the \textit{geometric layer}, which is smoothly distributed on the actual surface and normals are available. 
While the Gaussian positions with additional offset
\begin{equation}
    y_i=R_i(\theta)(\bar x^{\rm smpl}_i+\Delta\bar x_i+\Delta\bar y_i)+t_i(\theta)
\end{equation}
are referred to as the \textit{rendering layer}, which will be used for final rendering in this stage. 
To constrain the body layer lies beneath the clothing one, we employ a collision loss between the body geometric layer and the clothing geometric layer:
\begin{equation}
    \mathcal L_{\rm coll}=\frac{1}{N_{\rm cloth}}\sum_i\max(\epsilon-d_i, 0)^2,
\end{equation}
where $d_i=(\bar x^{\rm cloth}_i-\bar x^{\rm body}_i)\cdot \bar n^{\rm body}_i$, $\epsilon$ is a distance threshold and $i$ ranges over all valid pixels in the clothing Gaussian map. Note that simply handling collision between the body geometric layer and the clothing geometric layer does not ensure their corresponding rendering layers are separate. We therefore also constrain the rendering layer to stay close to the geometric layer:
\begin{eqnarray}
    \mathcal L_{\rm layer}&=&\frac{1}{N_{\rm body}}\sum_i\max(\|\Delta\bar y^{\rm body}_i\|_2-\epsilon/2,0)^2\nonumber\\
    &&+\frac{1}{N_{\rm cloth}}\sum_{i'}\max(\|\Delta\bar y^{\rm cloth}_{i'}\|_2-\epsilon/2,0)^2.
\end{eqnarray}
Intuitively, $\mathcal L_{\rm coll}$ encourages the clothing geometric layer and the body geometric layers to be at least $\epsilon$ apart while $\mathcal L_{\rm cloth}$ encourages the rendering layers to be at most $\epsilon/2$ apart from their corresponding geometric layers. Consequently, the two losses together pull both rendering layers apart to avoid collisions.

\subsubsection{Geometric Supervision from Reconstructions}
Recall that in the single-layer stage, we can already obtain segmented point clouds of clothing for each frame. 
Since using only image supervision to track the clothing boundaries is difficult, we directly supervise the clothing movement in this stage by enforcing a Chamfer distance loss between the clothing geometric layer and the segmented clothing reconstruction:
\begin{equation}
    \mathcal L_{\rm cd}={\rm ChamferDist}(\{x_i^{\rm cloth}\}_i,\{x_i^{\rm recon}\}_i),
\end{equation}
where $\{x_i^{\rm cloth}\}_i$ denotes the point cloud of the clothing geometric layer, and $\{x_i^{\rm recon}\}_i$ is the point cloud reconstructed from the first single-layer stage.

\subsubsection{Segmentation Loss} Since we use a fixed point set to represent clothing in this stage, we do not predict Gaussian labels $p^{\rm body}_i$ and $p^{\rm cloth}_i$. Instead, they are fixed as $(p^{\rm body}_i,p^{\rm cloth}_i)=(1,0)$ and $(p^{\rm body}_i,p^{\rm cloth}_i)=(0,1)$ for each Gaussian associated to the body and the clothing, respectively. Only $\mathcal L_{\rm label}$ in $\mathcal L_{\rm seg}$ is preserved in this stage.
We maintain the segmentation loss to avoid ambiguity, otherwise, if the clothing is optimized to be transparent, the body will take the colors of the clothing.

Finally, the total training loss in this stage is:
\begin{equation}
    \mathcal L=\mathcal L_{\rm render}+\mathcal L_{\rm geom}+\lambda_{\rm label}\mathcal L_{\rm label}+\lambda_{\rm coll}\mathcal L_{\rm coll}+\lambda_{\rm layer}\mathcal L_{\rm layer}+\lambda_{\rm cd}\mathcal L_{\rm cd}.
\end{equation}

\begin{figure*}
    \centering
    \includegraphics[width=\linewidth]{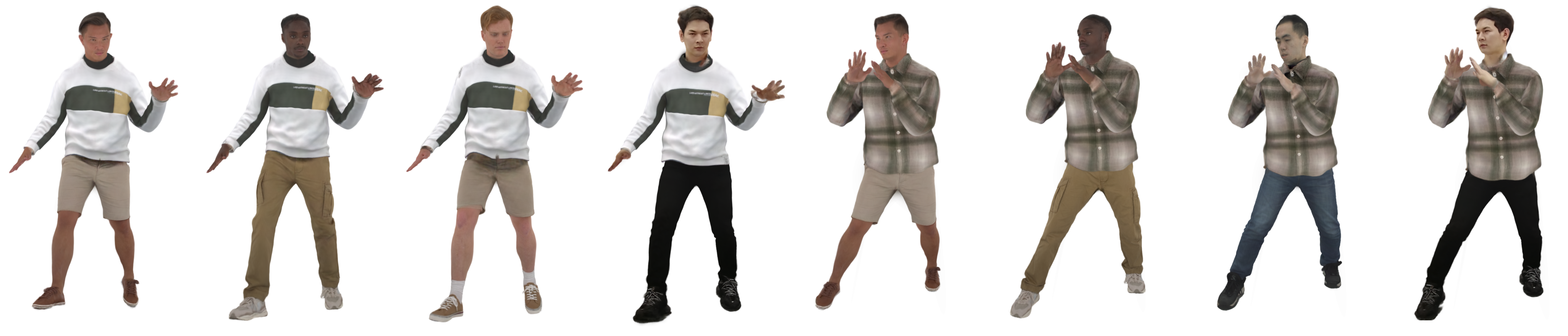}
    \caption{Our method enables animatable clothing transfer, and each row illustrates animation results with the same upper garment but different identities.}
    \label{fig:novel1x8}
\end{figure*}
\begin{figure}
    \centering
    \includegraphics[width=0.9\linewidth]{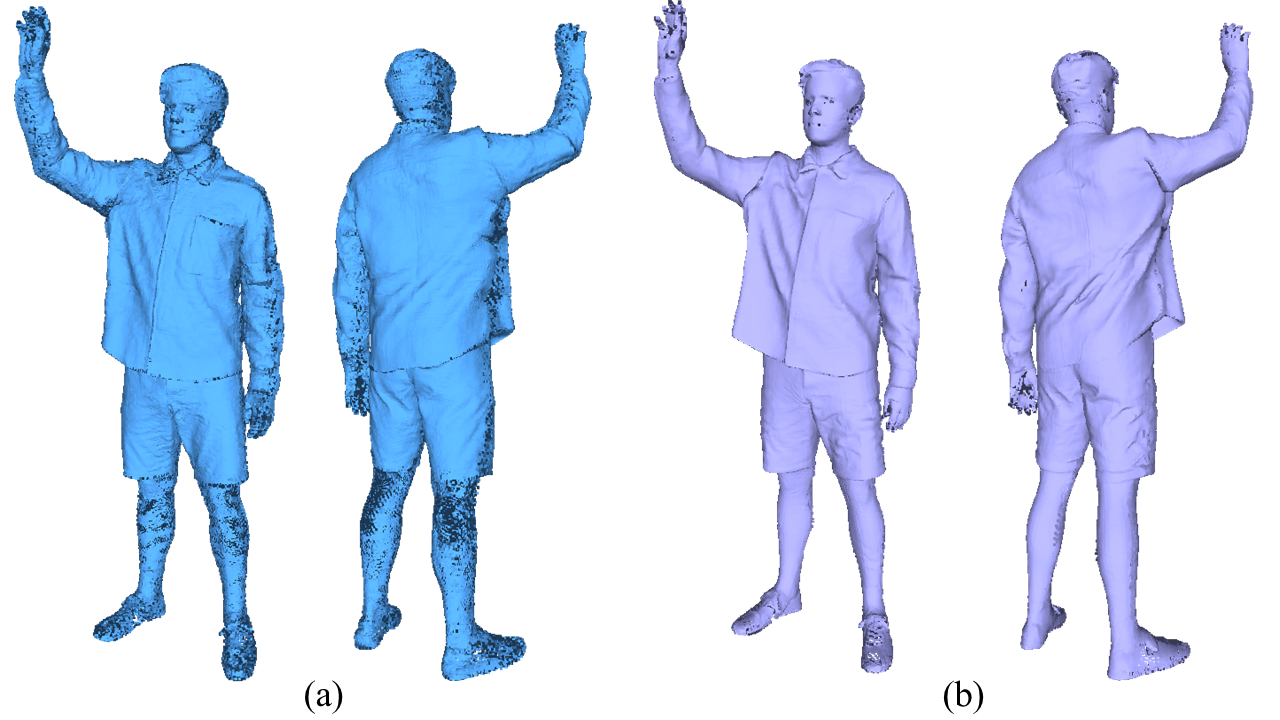}
    \caption{Geometric reconstruction results of the baseline method (a) and our method (b). Each Gaussian is shaded by its normal and rendered as a point. Our method exhibits better reconstruction quality, validating the effectiveness of our proposed geometric constraints.}
    \label{fig:recon}
\end{figure}

\subsection{Animatable Clothing Transfer and Collision Handling}
\label{sec:subsec:collision}

Once the training of both stages finishes, we obtain two models (body and clothing) for each captured subject. 
At test time, we can animate the subject using a novel pose sequence. 
Unfortunately, despite the effort during training to avoid collision, results are not guaranteed to be collision-free for novel poses, especially if we want to dress a garment on a new body shape.
To resolve collision, we propose an additional \textit{Laplacian-based} post-processing step.

Given a novel pose $\theta$ and two trained layered models\footnote{$\mathcal F_A^{\rm cloth}$ denotes the submodel of $\mathcal F_A$ that outputs the clothing related Gaussians. Other notations are defined similarly.} $\mathcal F_A=(\mathcal F_A^{\rm cloth},\mathcal F_A^{\rm body})$ and $\mathcal F_B=(\mathcal F_B^{\rm cloth},\mathcal F_B^{\rm body})$, suppose we want to transfer the clothing of $A$ to $B$. Let us denote the posed positions of garment Gaussians (the geometric layer) outputted by $\mathcal F^{\rm cloth}_A$ by $x_A^{\rm cloth}\in\mathbb{R}^{N^{\rm cloth}_A\times3}$, and respectively, the posed body Gaussian positions and normals of $\mathcal F^{\rm body}_B$ by $x_B,\ n_B^{\rm body}\in\mathbb{R}^{N_B^{\rm body}\times3}$. We attempt to resolve the collisions that happens between $x_A^{\rm cloth}$ and $x_B^{\rm body}$.

The basic idea is to pull $x_A^{\rm cloth}$ away from $x_B^{\rm body}$ along $n_B$ but keeping its graph Laplacian invariant. Let $L_A^{\rm cloth}$ denotes the graph Laplacian of $x_A^{\rm cloth}$ and $b=L_A^{\rm cloth}x_A^{\rm cloth}$ their Laplacian coordinates. Here, we consider two points in $x_A^{\rm cloth}$ as neighbors if they are neighbors in the Gaussian maps, or if they are a pair of pixels in the front and back Gaussian maps that should be stitched together. Note that since $A$ and $B$ may correspond to different body shapes, $x_A^{\rm cloth}$, $x_B^{\rm body}$ may severely collide with each other. To provide a collision-free initial guess of new positions of $x_A^{\rm cloth}$, we compute:
\begin{equation}
    \bar\xi=(\bar x_A^{\rm cloth}-U_A^{\rm cloth}\bar x_A^{\rm body})+U_A^{\rm cloth}\bar x_B^{\rm body}.\label{eq:coll-init-guess}
\end{equation}
Here, $U_A^{\rm cloth}$ is a mask matrix that selects the subset of garment Gaussians as described in the beginning of Sec.~\ref{subsec:multi-layer}. In other words, $\bar\xi$ is the canonical-space point cloud deformed from body B, with the relative displacements of $x_A^{\rm cloth}$ w.r.t. $x_A^{\rm body}$. We then apply LBS with the joint transformations of $B$ to pose $\bar\xi$ as $\xi$. We then solve (in the least squares sense)
\begin{equation}
    \begin{bmatrix}
        L_A^{\rm cloth}\\
        \alpha I\\
        \alpha I
    \end{bmatrix}\eta
    =\begin{bmatrix}
        b\\
        \alpha x_A^{\rm cloth}\\
        \alpha \xi
    \end{bmatrix}.
\end{equation}
In plain words, the solution $\eta$ tries to keep the Laplacian coordinates invariant, while remaining close to the original points $x_A^{\rm cloth}$ and the collision-free initial guess $\xi$.

We now update the current value of $x_A^{\rm cloth}$ to $\eta$. Note that $x_A^{\rm cloth}$ is still not yet collision-free with $x_B^{\rm body}$. To further refine the results, for each point in $x^{\rm cloth}_A$, we find its nearest neighbor in $\bar x^{\rm body}_B$. Let these nearest neighbors and their normals be denoted by $x_B^{\rm body,nn}$ and $n_B^{\rm body,nn}$. We compute the distance from $x_A^{\rm cloth}$ to $x_B^{\rm body,nn}$ as
\begin{equation}
    d_{\rm cloth2body}=(x_A^{\rm cloth}-x_B^{\rm body,nn})\cdot n_B^{\rm body,nn}\in\mathbb{R}^{N_A^{\rm cloth}}.
\end{equation}
The dot denotes point-wise inner product. We then set
\begin{equation}
\xi=x^{\rm cloth}_A+{\rm clip}(\epsilon-d_{\rm cloth2body})|_{{\rm min}=0}^{{\rm max}=\delta}\odot\bar n^{\rm body}_i.
\end{equation}
Here, $\odot$ means point-wise scalar-vector multiplication. Another Laplacian system is solved (in the least squares sense) to obtain update $x_A^{\rm cloth}$:
\begin{equation}
    \begin{bmatrix}
        L_A^{\rm cloth}\\
        \alpha I
    \end{bmatrix}\eta
    =\begin{bmatrix}
        b\\
        \alpha \xi
    \end{bmatrix}.
\end{equation}
In plain words, we first find a new guess $\xi$ by moving $x_A^{\rm cloth}$ away from $x_B^{\rm body,nn}$ along the normals, but during each update, the moving distance is limited by $\delta$ and $\epsilon-d_{\rm cloth2body}$. The updated $x_A^{\rm cloth}$ is then obtained by solving a Laplacian system involving $\xi$. This process is repeated five times.

After $x^{\rm cloth}_A$ has been optimized as above, its corresponding rendering layer is obtained by adding the posed offsets of $\Delta\bar y^{\rm cloth}_A$. This will be used for the final rendering. We remark that collision handling is only used at test time.

\begin{table}%
\caption{Quantitative evluation of rendering quality.}
\label{tab:one}
\footnotesize
\begin{center}
\begin{tabular}{llcccc}
  \toprule
  Subject & Method & PSNR$\uparrow$ & SSIM$\uparrow$ & LPIPS$\downarrow$ & FID$\downarrow$ \\
  \midrule
  \multirow{2}{*}{A02} & \citet{li2023animatable} & 28.2567 & 0.9603 & 0.0396 & \textbf{25.3897} \\
   & Ours & \textbf{28.5727} & \textbf{0.9630} & \textbf{0.0380} & 26.1757 \\
  \hline
  \multirow{2}{*}{A05} & \citet{li2023animatable} & \textbf{28.8049} & \textbf{0.9595} & \textbf{0.0430} & \textbf{33.4961}\\
   & Ours & 27.3584 & 0.9478 & 0.0503 & 39.3849 \\
  \hline
  \multirow{2}{*}{A08} & \citet{li2023animatable} & 29.0913 & 0.9565 & 0.0377 & 22.0500\\
   & Ours & \textbf{29.2883} & \textbf{0.9583} & \textbf{0.0367} & \textbf{21.4906}\\
  \bottomrule
\end{tabular}
\end{center}
\end{table}%

\section{Experiments}

In this section we present our main results on geometric reconstruction and animated clothing transfer. Due to page limits, please refer to our supplementary document and video for extended evaluations and discussions.

\subsection{Dataset and Training}

We train our model on sequences from two datasets: one sequence from the AvatarRex dataset and three sequences (A02, A05, A08) from the ActorsHQ dataset~\cite{isik2023humanrf}. Additionally, we also capture a new multi-view sequence of a person in a white shirt dancing to train our model. Our training/evaluation setup and model architectures are mostly the same as \citet{li2023animatable}. The differences are detailed in the supplementary material.

\subsection{Rendering and Reconstruction Quality}

Since there is currently no open-source work that models layered animatable avatars from multi-view videos, we mainly compare with our baseline: Animatable Gaussians~\cite{li2023animatable}, which models body and clothing as a whole and do not attempt geometric reconstruction. We quantitatively evaluate the rendering quality using PSNR, SSIM~\cite{wang2004image}, perceptual loss (LPIPS) \cite{zhang2018unreasonable} and FID~\cite{heusel2017gans}.

To evaluate the effectiveness of our geometric constraints, we compare the geometry of the Gaussians reconstructed by the baseline method \citep{li2023animatable} (single-layer without geometric constraints) and ours. Note that the baseline method does not provide normals. Thus, we compute normals using the method illustrated in Fig.~\ref{fig:normalcompute}. Fig.~\ref{fig:recon} shows geometric results of both methods, where Gaussians are rendered as ordinary point clouds with normals. For the baseline method, the messy shading near the leg area indicates incorrect normal orientation. This suggests their reconstructed Gaussians do not lie on the actual geometric surface. On the other hand, our method produces clean point cloud reconstructions using Gaussians.

\subsection{Animating Layered Avatars and Clothing Transfer}

\begin{figure*}
    \centering
    \includegraphics[width=\linewidth]{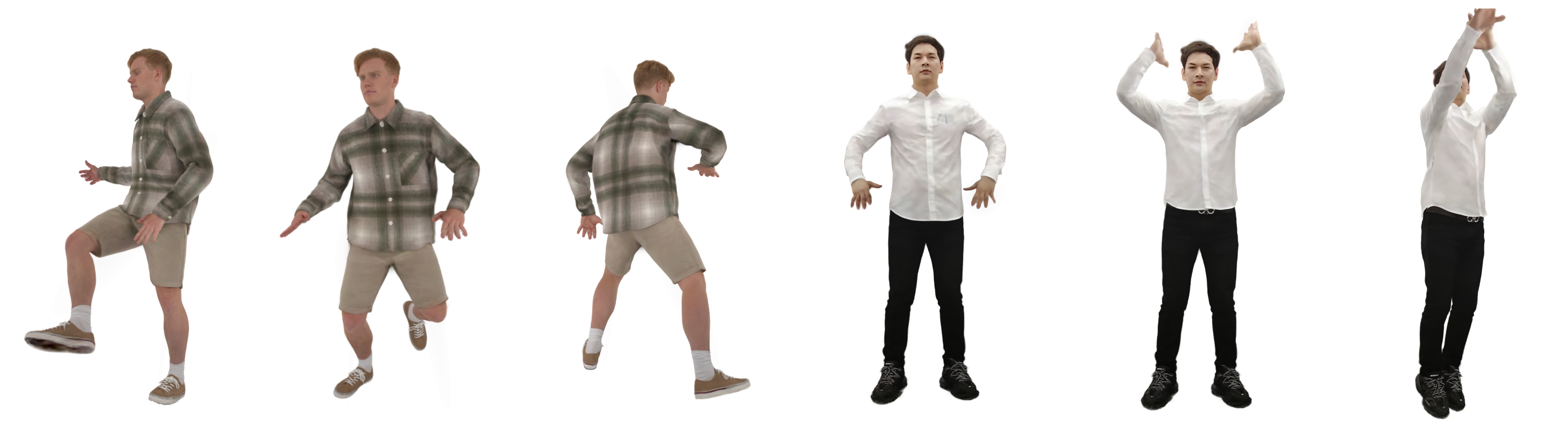}
    \caption{Qualitatively results of novel pose animation. Our layered representation can model tangential motions between body and clothing, e.g., when the T-shirt is lifted and the belt is revealed.}
    \label{fig:novelpose}
\end{figure*}

\begin{figure*}
    \centering
    \includegraphics[width=\linewidth]{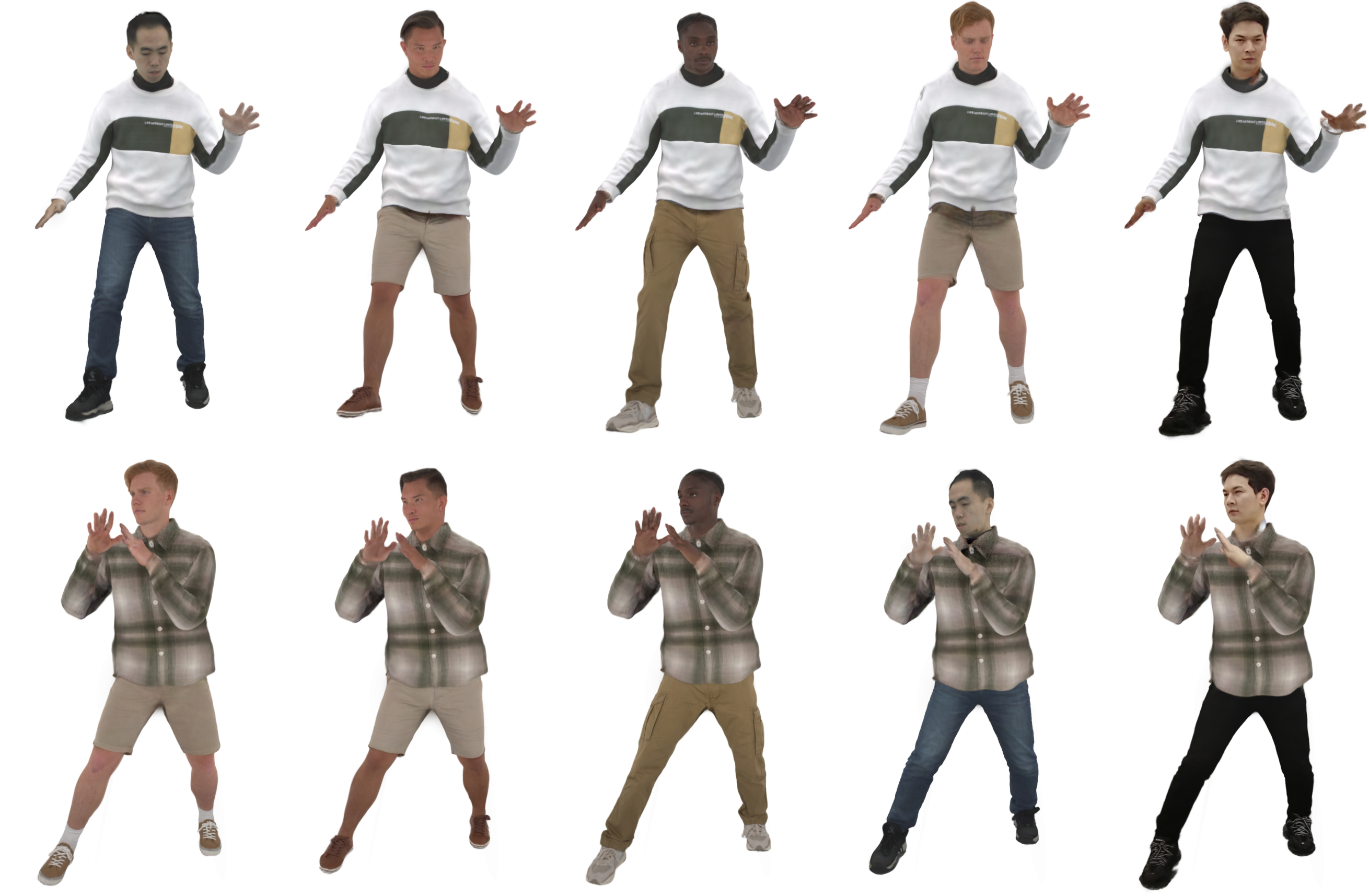}
    \caption{Our method enables animatable clothing transfer, and each row illustrates animation results with the same upper garment but different identities.}
    \label{fig:novel}
\end{figure*}

Given trained LayGA models of different subjects, we can animate one subject or use the garment model of $A$ and the body model of $B$ to generate a mixed avatar. Fig.~\ref{fig:novelpose} shows novel pose animation results. Note that our layered representation can model tangential motions of clothing. Fig.~\ref{fig:novel1x8} (Fig.~\ref{fig:novel} shows an enlarged version) exhibits clothing transfer results, where the leftmost subject in each row provides the clothing model while others are the target shapes. Thanks to our multi-layer design and collision resolving strategies in Sec.~\ref{sec:subsec:collision}, our method is capable of dressing a given garment to different body shapes and generating photorealistic renderings.

\section{Discussion}

In this paper, we present Layered Gaussian Avatars (LayGA) for animatable clothing transfer.
We propose two-stage training involving single-layer reconstruction and multi-layer fitting.
In the single-layer reconstruction stage, we propose a series of geometric constraints for surface reconstruction and segmentation.
In the multi-layer fitting stage, we train two separate models to represent the body and clothing, thus enabling clothing transfer across different identities.
Overall, our method outperforms the state-of-the-art baseline \citep{li2023animatable} and realizes photorealistic virtual try-on. Moreover, our geometrically constrained Gaussian rendering scheme, if considered as a stand-alone method, can also be used for multi-view geometry reconstruction of humans. 

\begin{figure*}[t]
    \centering
    \includegraphics[width=0.8\linewidth]{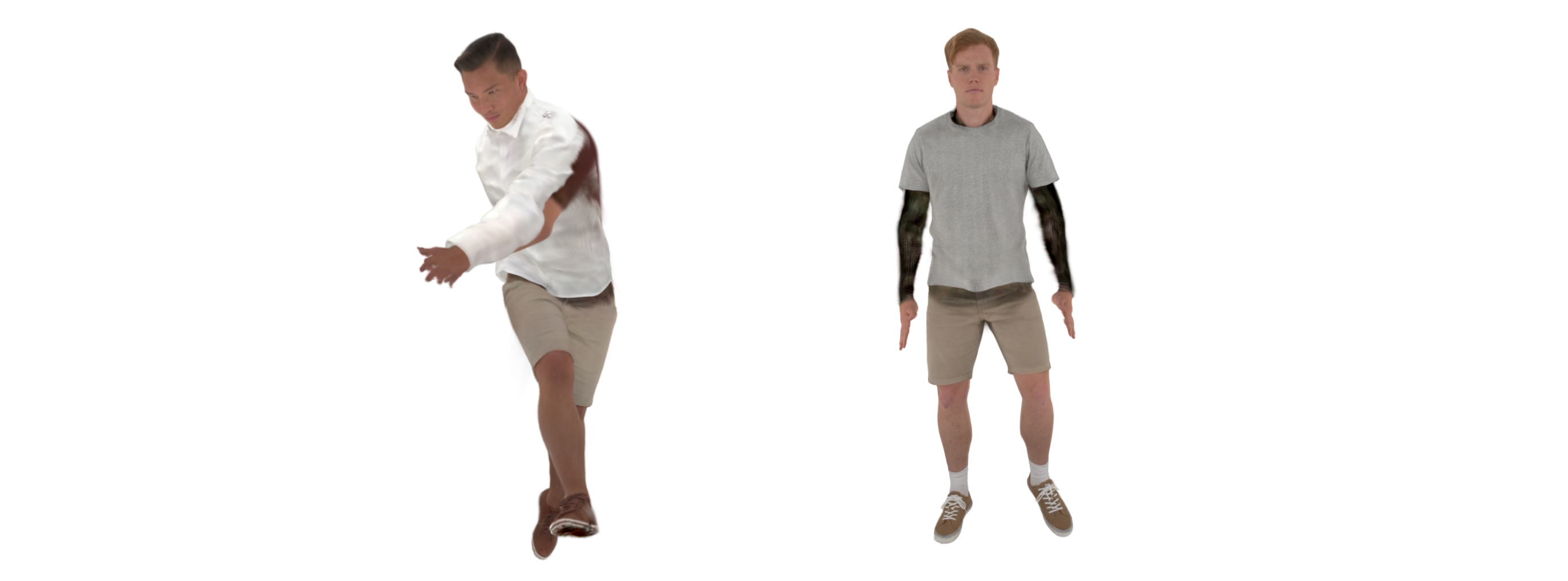}
    \caption{Failure Cases. Left: Transferring tight clothing to a larger body shape causes penetration that cannot be resolved by collision handling. Right: Body model trained with long-sleeves have undefined colors, opacities, etc., in occluded areas (e.g., arms). If a short-sleeve T-shirt is dressed to it, these areas would not produce correct rendering results.}
    \label{fig:failure}
\end{figure*}

Despite its good performance, our method still suffers from limitations: (a) If the source garment is tight and the target body shape is large, then our collision handling method may still fail (Fig.~\ref{fig:failure} left). (b) If a short-sleeve T-shirt is transferred to a body model wearing long sleeves, the arm part would not produce correct rendering results (Fig.~\ref{fig:failure} right). This is because the arm part of the target shape is occluded during training, with no supervision forcing it to take on skin colors. (c) The approach is not designed to simulate the actual physics during clothing transfer, and the results may not be physically realistic when transferring between very different body sizes. We leave these as future work.

\begin{acks}
The work is supported by National Key R\&D Program of China (2022YFF0902200), the National Science Foundation of China (NSFC) under Grant Number 62125107 and the Postdoctoral Fellowship Program of China Postdoctoral Science Foundation under Grant Number GZC20231304.
\end{acks}

\bibliographystyle{ACM-Reference-Format}
\bibliography{reference}


\end{document}